\documentclass[letterpaper, 10 pt, conference]{ieeeconf}
\usepackage{amsmath}
\usepackage{mathtools}
\usepackage{algorithm}
\usepackage{setspace}
\usepackage[noend]{algpseudocode}
\usepackage{hyperref}
\usepackage{bm}
\usepackage{hhline}
\usepackage{multirow}

\usepackage{siunitx}

\usepackage{subcaption}
\usepackage{kantlipsum} 
\usepackage{mwe} 
\usepackage{booktabs}

\DeclareCaptionLabelSeparator{periodspace}{.\quad}
\IEEEoverridecommandlockouts                              

\title{\LARGE \bf
GPU Accelerated Voxel Grid Generation for fast MAV Exploration
}

\author{Charbel Toumieh and Alain Lambert
\thanks{The two authors are with the Universit\'e Paris-Saclay, CNRS, Laboratoire Interdisciplinaire des Sciences du Numérique, 91405, Orsay, France \url{https://www.lisn.upsaclay.fr}}%
}
\begin{document}

\maketitle
\thispagestyle{empty}
\pagestyle{empty}

\begin{abstract}
Voxel grids are a minimal and efficient environment representation that is used for robot motion planning in numerous tasks. Many state-of-the-art planning algorithms use voxel grids composed of free, occupied and unknown voxels. In this paper we propose a new GPU accelerated algorithm for partitioning the space into a voxel grid with occupied, free and unknown voxels. The proposed approach is low latency and suitable for high speed navigation.
\end{abstract}


\section{INTRODUCTION}
Many sensors (RGB-D cameras, stereo-matching ...) output dense pointclouds as measurements and need to be processed and turned into an environment model/representation for motion planning.
Fast 3D environment modelling is crucial for real-time high speed motion planning and exploration.

Many state-of-the art techniques use voxel grids for planning \cite{ryll2019efficient} \cite{tordesillas2020faster} \cite{tordesillas2018real} \cite{toumieh2020planning} \cite{toumieh2022multi} \cite{toumieh2022near} \cite{toumieh2022dyn} \cite{toumieh2022mace}. Some of them use the grid for Safe Corridor generation \cite{liu2017planning} \cite{toumieh2022convex} \cite{toumieh2022shapeaware}, while others use it for direct collision checking. They need the grid to be partitioned into occupied, free and unknown voxels. The fast generation of such grids is the main objective of this paper.

We will first present the related work as well as our main contributions. Then, we will outline the different steps of the method and show the simulation results. Finally, we will compare it with \textbf{mit-acl-mapping}\footnote{https://gitlab.com/mit-acl/lab/acl-mapping} used in \cite{ryll2019efficient}. Our method is implemented on CPU and GPU and we will discuss the performance difference throughout the paper.

\subsection{Related Work}
Numerous 3D space representations exist such as signed distance fields \cite{oleynikova2017voxblox} \cite{han2019fiesta}, octrees \cite{hornung2013octomap} \cite{duberg2020ufomap}, and voxel grids \cite{ryll2019efficient}.

Signed distance fields grids include in every voxel information about the distance and gradient against obstacles and is essential for gradient-based planning methods. The advantage of voxel grids over signed distance fields is processing time, since they don't need to calculate the distance of every voxel to the closest obstacle.

Octrees are a tree data structure where each node has eight children. Octree based representations such as Octomap \cite{hornung2013octomap} are an efficient probabilistic 3D representation of the environment. The advantage of voxel grids over octrees is the constant voxel access (read/write) time.

 In \cite{hermann2014unified}, the authors use General Purpose 
 Graphics Processing Unit (GPGPU) to populate the voxel grid. The authors raycast every measurement point in a Brensenham fashion \cite{bresenham1965algorithm} and update the voxels according to the sensor model.
 
 In \cite{ryll2019efficient} authors use \cite{bresenham1965algorithm} to trace every pixel of the image and free the traversed voxels between the camera center and the pixel depth. 
  This approach leads to a high computational cost and become intractable for medium/high resolution RGB-D cameras.

\subsection{Contribution}
In this work we propose a new implementation of a voxelization algorithm on the GPU that tries to minimize computation time as much as possible. It takes as input a dense point cloud and outputs a local voxel grid centered at the robot with occupied, free, and unknown cells. However we don't use a probabilistic approach as this reduces performance of the GPU in our implementation (requires atomic operations). 

\section{Nomenclature}
We define the various abbreviations and variables used throughout this paper in table \ref{table_nomenclature}. All distances and coordinates are in meters except for voxel coordinates which are integers.
\begin{table}[ht]
\caption{Nomenclature}
\label{table_nomenclature}
\begin{center}
\begin{tabular}{c l}
\hline
$loc\_grid$ & local voxel grid \\
$ms\_grid$ & measurement voxel grid \\
$grid\_size_x$ & grid size in the $x$ direction\\
$grid\_size_y$ & grid size in the $y$ direction\\
$grid\_size_z$ & grid size in the $z$ direction\\
$vox\_size$ & voxel cube side length\\
$fov_x$ & camera field of view in the $x$ direction\\
$fov_y$ & camera field of view in the $y$ direction\\
$depth$ & IR depth sensor range\\
$vox\_depth$ & the $depth$ in number of voxels\\
$vox\_width$ & voxels covered by $fov_x$ at $depth$\\
$vox\_height$ & voxels covered by $fov_y$ at $depth$\\
$x_i$ & voxel $x$ coordinate in the grid frame\\
$y_i$ & voxel $y$ coordinate in the grid frame\\
$z_i$ & voxel $z$ coordinate in the grid frame\\
$\boldsymbol{T}^w_c$ & transform from camera to world frame\\
$\boldsymbol{T}^v_w$ & transform from world to grid frame\\
$\boldsymbol{p}^c_o$ & obstacle point in the camera frame\\
$\boldsymbol{p}^v_o$ & obstacle point in the voxel grid frame\\
$\boldsymbol{p}^v_c$ & camera position in the voxel grid frame\\
$vox\_inf$ & number of voxels to inflate\\
$ray\_dir$ & direction of the ray to trace\\
$ray\_start$ & starting point of the ray to trace\\
$max\_dist$ & maximum ray traversal distance\\
\hline
\end{tabular}
\end{center}
\end{table}


\begin{figure}
\centering
\includegraphics[trim={0cm 0cm 0cm 0cm},clip,width=1\linewidth]{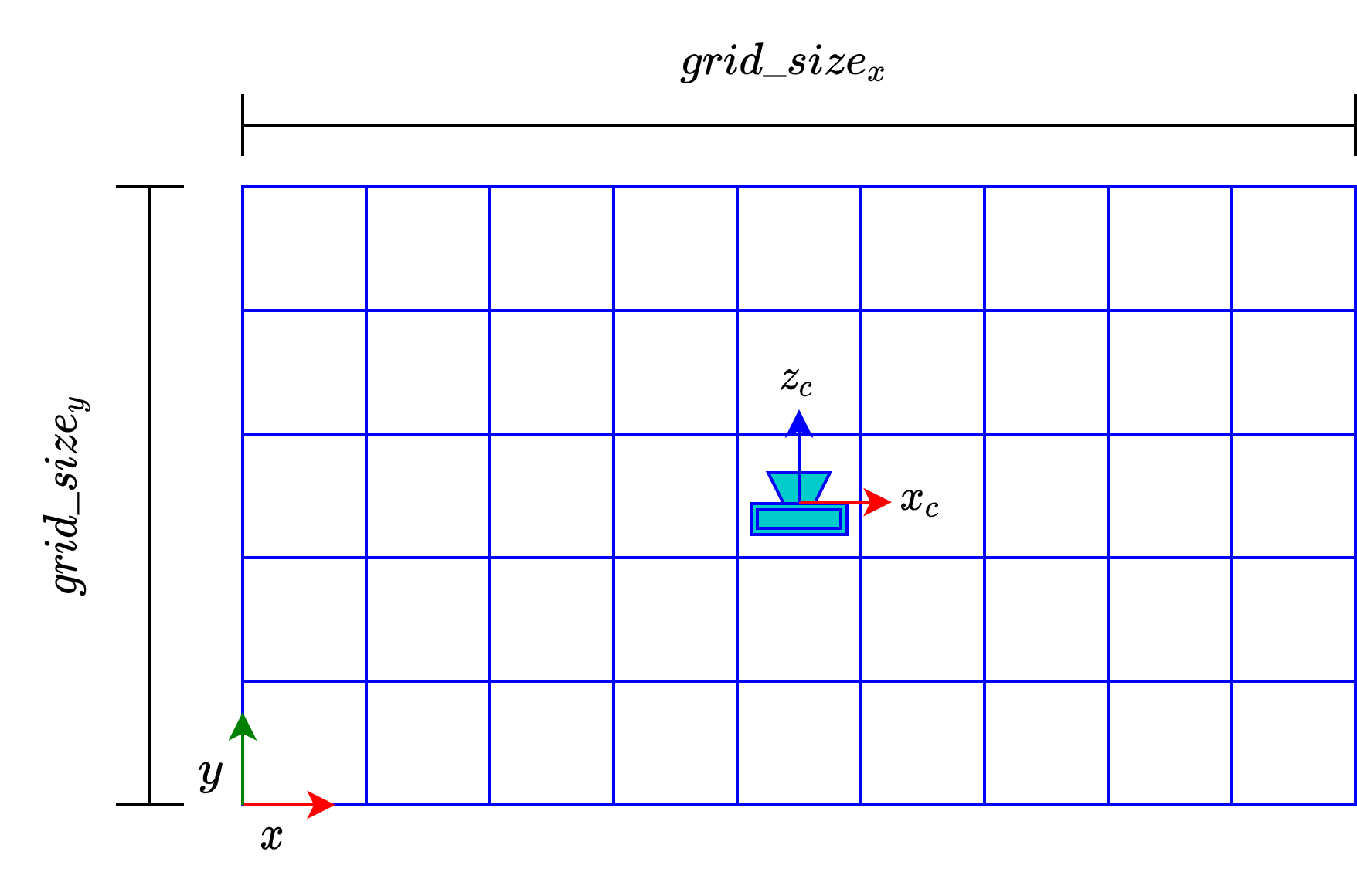}
\caption{We show a 2D example of the grid with the origin set such as the robot/camera position is at the center of the grid.}
\label{fig:frame}
\end{figure}

\section{GPU architecture}
In this section we briefly describe the GPU architecture/nomenclature and how it executes instructions. This will help make sense of the performance benchmarks in the simulation section. We will describe NVIDIA's Turing architecture which our GPU (RTX 2060) is built on \cite{cuda_c}.

The basic building block of a GPU is the Streaming Multiprocessor or SM. Each SM in the Turing architecture contains 64 cores (2 groups of 32 cores) and has its own L1 shared memory. In every group of 32 cores, all cores execute the same instructions simultaneously, but with different data (Single Instruction Multiple Threads - SIMT). Thus we get 32 threads (called a warp) all doing the same thing at the same time.

The warps (groups of 32 threads) are grouped into blocks and every block can be executed on one SM only. A group of blocks is called a grid.

Throughout the paper we will measure the efficiency of our implementations through active threads per warp (how many of the 32 cores are active - the higher the better), and the number of warp cycles per executed instruction (the lower the better).

Active threads per warp also depend on predication when we have \textbf{if} - \textbf{else} conditional statements. With predication, the GPU evaluates both sides of the conditional statement and then discards one of the results, based on the value of the boolean branch condition in each thread. Predication is in general effective for small branches.
 
The GPU code is written in CUDA \cite{nickolls2008scalable}.
\section{The method}
The objective is to generate a voxel grid representation of the world around the robot.
The method takes a dense point cloud (stereo matching) or a depth image (IR depth sensor) in the camera frame,  and populates a voxel grid with occupied, free and unknown cells.

The method consists into updating a sliding local voxel grid centered around the robot's position with a measurement grid that represents the latest measurement. It is divided into 5 steps (functional blocks):

\begin{enumerate}
    \item Voxel grid setup
    \item Populate occupied voxels.
    \item Ray-trace to free voxels in camera field of view.
    \item Update the local voxel grid with the measurement voxel grid.
    \item Shift the local grid to be centered at the robot's position.
\end{enumerate}

\begin{figure*}
\centering
\includegraphics[trim={0cm 0cm 0cm 0cm},clip,width=0.75\linewidth]{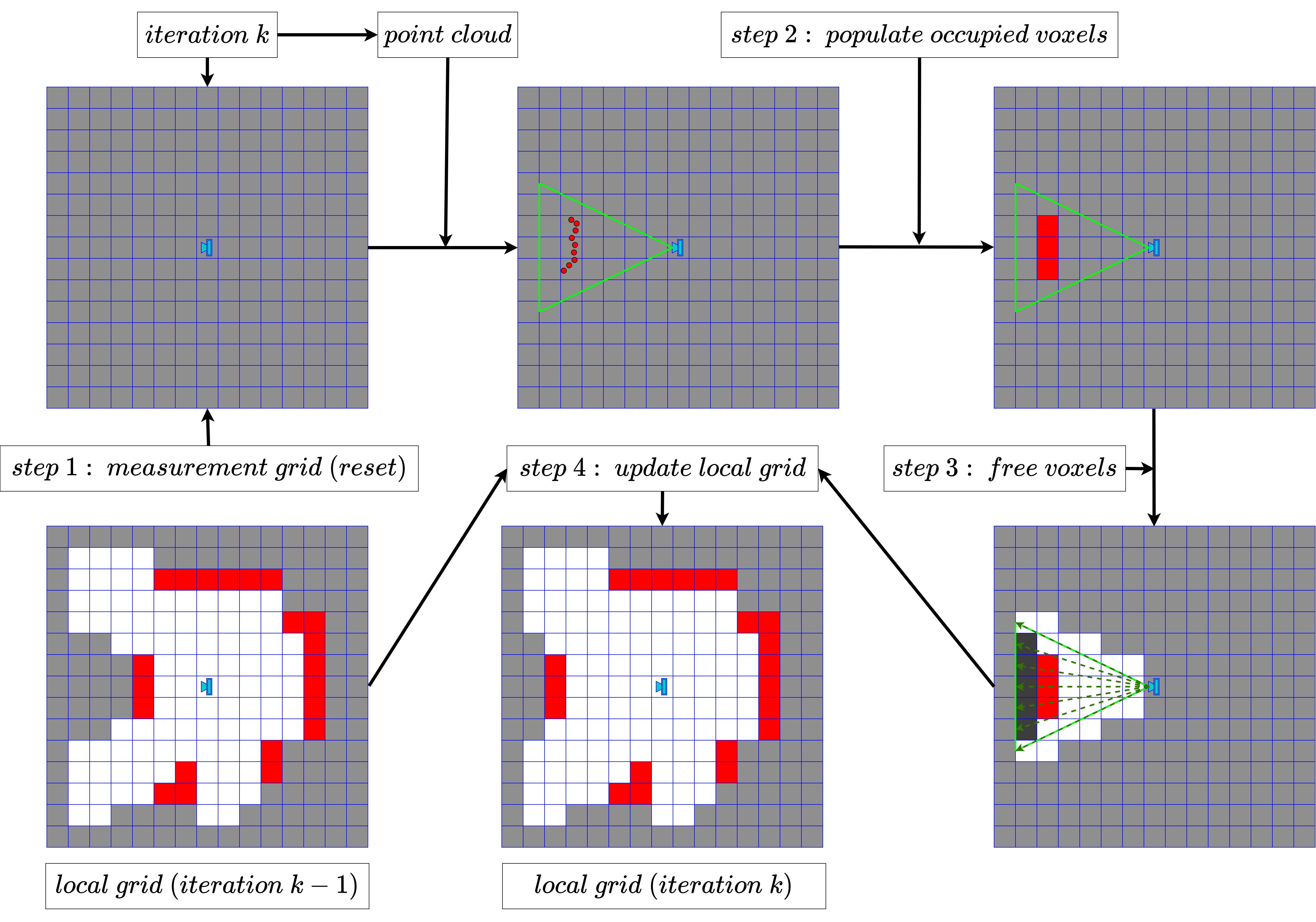}
\caption{We show the execution pipeline at iteration $k$ of steps 1-4 after receiving a measurement (point cloud). In the presented case, an dynamic obstacle moves away from the camera while the robot stays static. \textbf{Free} voxels are \textbf{white}, \textbf{occupied} are \textbf{red}, \textbf{unknown} are \textbf{grey}, \textbf{unknown and traced} are \textbf{black}. The camera field of view is shown in green and the point cloud as red circles.}
\label{fig:diag}
\end{figure*}

A diagram showing a global view of the execution pipeline of steps 1-4 is shown in Fig. \ref{fig:diag}.

In the voxel grid setup, we initialize the local voxel grid and measurement voxel grid. The local grid is only initialized once at the start of the algorithm and is updated at every measurement with the measurement voxel grid that is reset before each measurement.
We then set the voxels that contain obstacles to occupied in the measurement voxel grid (populate occupied voxels).

An efficient ray-tracing method is used in the third step (Ray-trace to free voxels in camera field of view). Instead of tracing each pixel we leverage the voxel grid structure to reduce considerably the number of rays to trace which accelerates the overall processing time. It is inspired by \cite{Yguel2006EfficientGC} and \cite{oleynikova2017voxblox} who uses ray bundling.

We then update the local voxel grid with the measurement one which we populated with the occupied voxels and freed its voxels with ray-tracing.
Finally we shift the local grid so that it is always centered at the robot position as the robot moves.
We assume a regular voxel grid, but the method can be generalized for irregular grids. Each step as well as its GPU implementation are discussed in what follows.

\subsection{Voxel grid setup}
We have 2 grids: the local voxel grid that is the representation of all previous measurements, and the measurement voxel grid that is the representation of the latest measurement. Both grids are of the same size. It is possible to use only one grid (the local grid). However, this would require clearing all the voxels in the camera field of view (using ray-tracing) before applying steps (2-4). We found empirically that this would result in a higher computation time then using 2 grids and merging them together (due to the high computation cost of ray-tracing).

The origin of the local voxel grid is initialized such as the initial robot/camera position is at the center of the grid. The orientation is the same as the ENU (East North Up) frame (Fig. \ref{fig:frame}). It is then shifted as the robot moves.
All voxels are initialized as \textbf{unknown}, and 
will be updated when the measurement voxel grid is merged with the local voxel grid. The voxels of the measurement voxel grid are set to \textbf{unknown} before every single measurement. They are then changed by steps 2 and 3.

\begin{figure}
\centering
\includegraphics[trim={0cm 0cm 0cm 0cm},clip,width=1\linewidth]{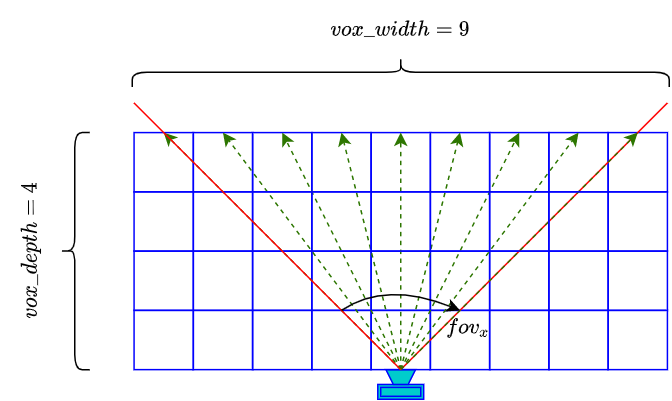}
\caption{We show a 2D example of the rays (in \textbf{green}) we trace using our method. The field of view of the camera is limited by the \textbf{red} lines.}
\label{fig:fov}
\end{figure}

\begin{figure}
\begin{subfigure}{0.24\textwidth}
\centering
\includegraphics[trim={0cm 0cm 0cm -1cm},clip,width=0.95\linewidth]{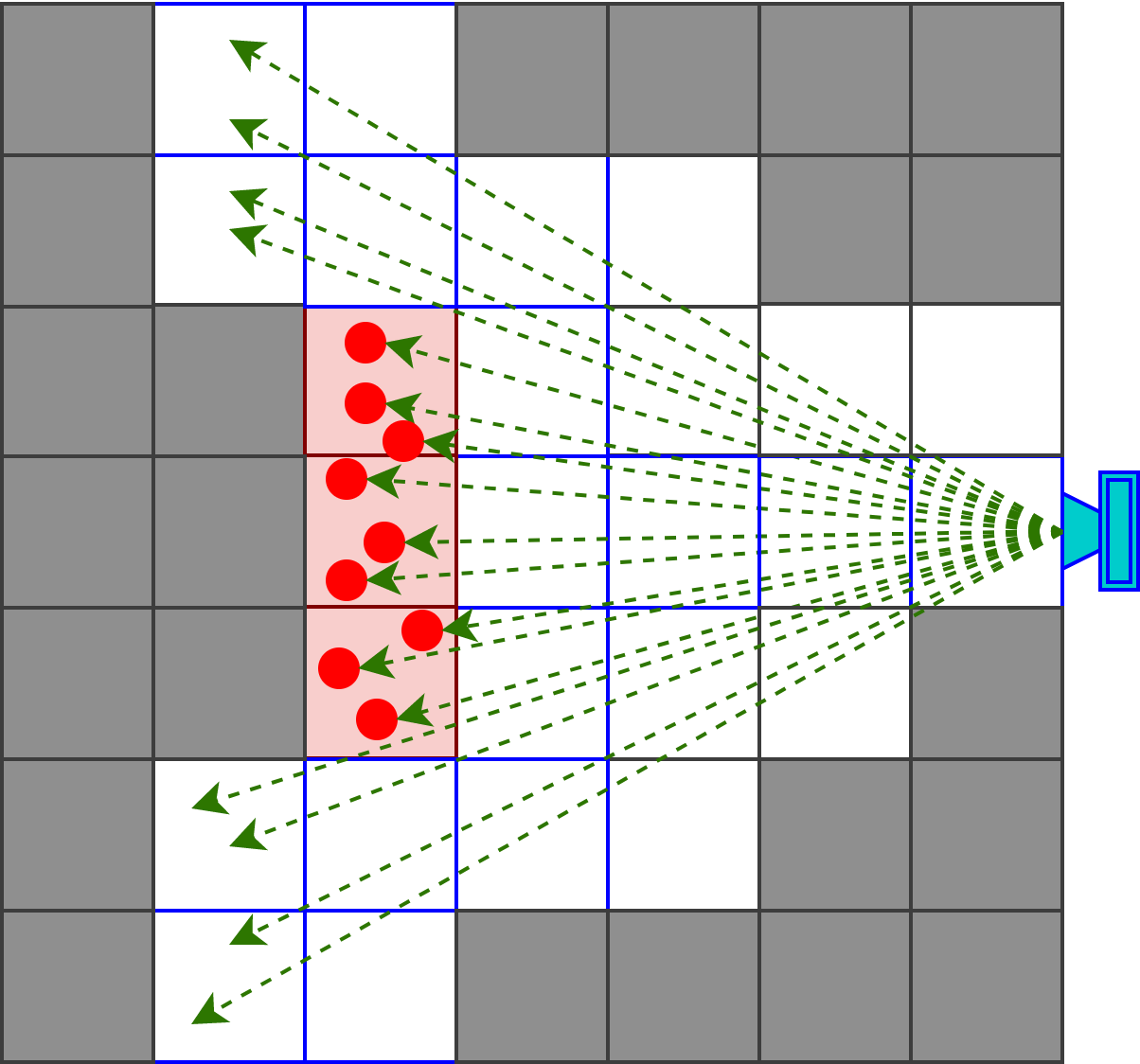}
\caption{Simple ray-tracing}
\label{fig:simple_ray}
\end{subfigure}
\begin{subfigure}{0.24\textwidth}
\centering
\includegraphics[trim={0cm 0cm 0cm -1cm},clip,width=0.95\linewidth]{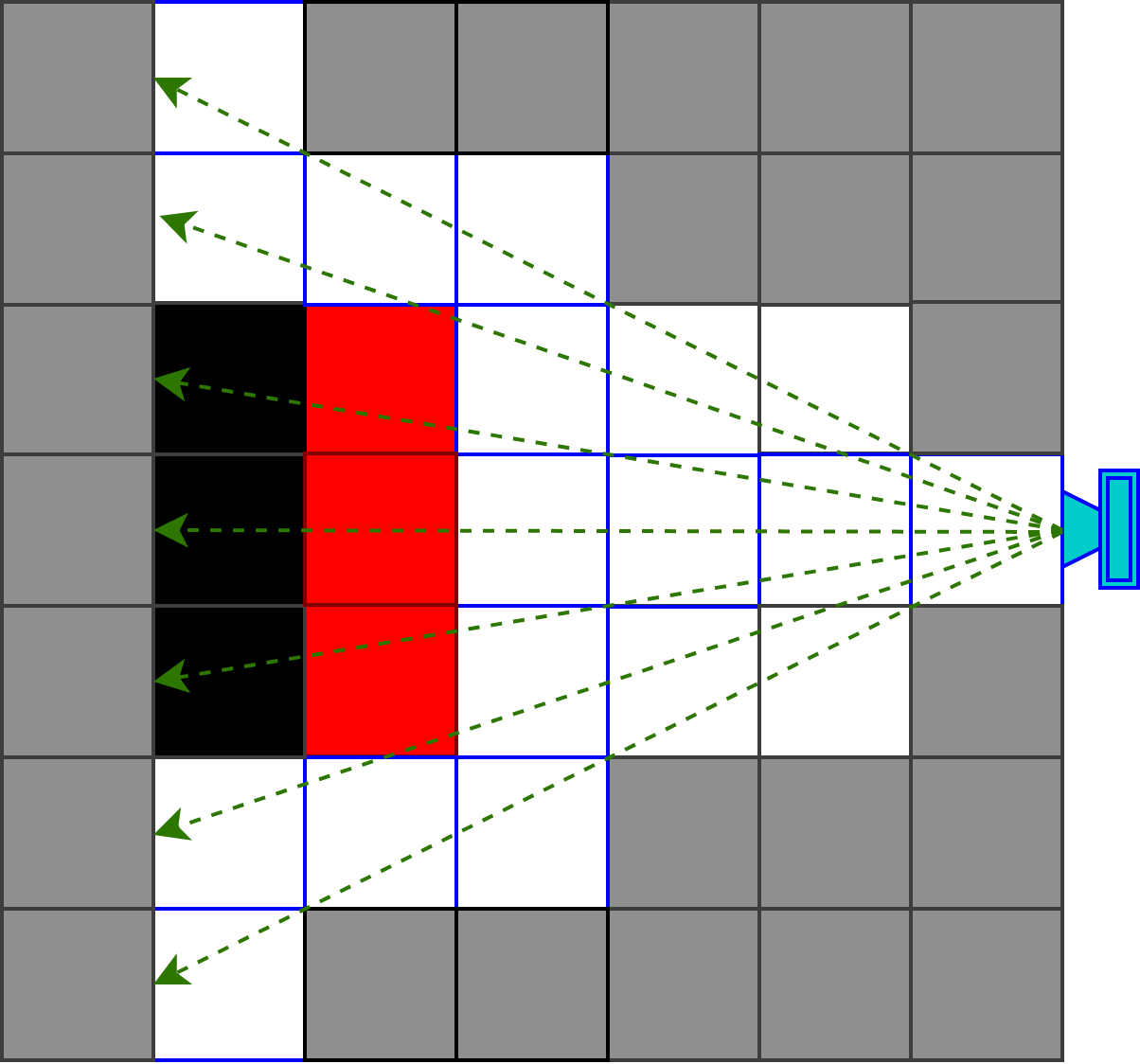}
\caption{Our ray-tracing}
\end{subfigure}
\caption{We show the \textbf{points} of a dense pointcloud in \textbf{red}, the \textbf{unknown} voxels in \textbf{grey}, \textbf{occupied} voxels in \textbf{red}, \textbf{free} voxels in \textbf{white} and \textbf{unknown and traced} in \textbf{black}. Our method (b) traces less rays with a close end result to (a). This speeds up computation time.}
\label{fig:comp_ray}
\end{figure}

\subsection{Populate occupied voxels}
In this step, we set all the voxels in the measurement grid that contain obstacle points to occupied using Alg. \ref{alg:occ_algo}.

This is done by first transforming the points from the camera frame to the voxel grid's origin (line 2), then dividing the coordinates by the voxel size to get the integer coordinates of the occupied voxel (lines 3-5). For every measurement/camera pose, the transformation matrix is the same: $\boldsymbol{T}^v_c =\boldsymbol{T}^v_w \boldsymbol{T}^w_c$. It is calculated on the CPU and passed as an argument to the GPU implementation.

Furthermore, certain planning methods require inflating the obstacles by a number of voxels $vox\_inf$. This can be implemented in this step (lines 6-9).

On the GPU, the frame transformation is done in parallel as well as setting the corresponding voxels to "occupied". If multiple threads in a warp want to concurrently set the same voxel to "occupied", only one of them succeeds while the others are discarded \cite{cuda_c}.

\begin{algorithm}
\caption{Populate occupied voxels}\label{alg:occ_algo}
\begin{algorithmic}[1]
\Function{AddOcc}{$\boldsymbol{T}^v_c,vox\_size,\boldsymbol{p}^c_o,ms\_grid,vox\_inf$}
\State   $\boldsymbol{p}^v_o = \boldsymbol{T}^v_c \boldsymbol{p}^c_o$
\State $x_i \gets \boldsymbol{p}^v_{o,x}/vox\_size$
\State $y_i \gets \boldsymbol{p}^v_{o,y}/vox\_size$
\State $z_i \gets \boldsymbol{p}^v_{o,z}/vox\_size$
\For{$i=x_i-vox\_inf$ \textbf{to} $x_i+vox\_inf$}
\For{$j=y_i-vox\_inf:$ \textbf{to} $y_i+vox\_inf$}
\For{$k=z_i-vox\_inf$ \textbf{to} $z_i+vox\_inf$}
\State $ms\_grid[i,j,k] \gets occupied$
\EndFor
\EndFor
\EndFor
\EndFunction
\end{algorithmic}
\end{algorithm}

\subsection{Ray-trace to free voxels}
In this step we free all the voxels in the measurement grid between the center of the camera and the occupied voxel since if any of them contained an obstacle, it would have been detected by the dense pointcloud representation.

Instead of tracing every point that is in a dense pointcloud, or every depth pixel in a depth image, we adopt another approach that significantly decreases the number of rays to trace (Alg. \ref{alg:unk_algo}). 

First we determine the $depth$ which we want to clear (e.g. range of the IR depth sensor) and calculate the number of voxels $vox\_ depth$ covered by it. This is done by dividing the $depth$ by the voxel size $vox\_size$.

Then, we determine $vox\_ width = 2*\tan(\frac{fov_x}{2})*vox\_ depth +1 $ with $fov_x$ the field/angle of view of the camera in the x direction (Fig. \ref{fig:fov}).

We also determine $vox\_ height = 2*\tan(\frac{fov_y}{2})*vox\_ depth +1 $ with $fov_y$ the field/angle of view angle of the camera in the y direction.

Note that $vox\_size$,  $vox\_ width$ and $vox\_ height$ are rounded to the closest integer.

Using \cite{amanatides1987fast},
 we trace all the rays who start from the camera center, with each ray having as direction one of the vectors going from the camera center to the voxels whose integer coordinates are:
\begin{gather*}
    z_i = vox\_ depth \\
    -\frac{(vox\_ height-1)}{2} \leq y_i \leq \frac{(vox\_ height-1)}{2} \\
    -\frac{(vox\_ width-1)}{2} \leq x_i \leq \frac{(vox\_ width-1)}{2}
\end{gather*}

We subtract 1 from $vox\_height$ and $vox\_width$ since by construction they are odd numbers (Fig. \ref{fig:fov}).
Since the camera frame is not always aligned with the voxel grid frame,
 we have to transform the ray direction vector from the camera to the voxel grid origin frame (line 4).

The voxels that the ray traverses before hitting an \textbf{occupied} voxel are set to free. All the voxels traced after hitting the occupied voxel are set to \textbf{unknown and traced} so they are differentiated from the \textbf{unknown} voxels outside the sensor field of view during the update (lines 8-14). Each ray is stopped when it traverses a predefined distance (e.g. $depth$ distance or $vox\_size\sqrt{x_i^2 + y_i^2 + z_i^2}$) Fig. \ref{fig:comp_ray}. 

This ray-tracing method assumes that all the obstacles in the camera field of view and within a certain depth are detected, which is ensured with RGB-D cameras or stereo-matching using state-of-the-art methods \cite{tankovich2020hitnet}.

On the GPU, we may have concurrent writes by some threads. This results in an undefined behavior on the GPU, as some voxels may have different values written to them concurrently by different rays. This only affects voxels on the borders of a region occluded by an obstacle, and may not affect the overall performance depending on the application.

\begin{algorithm}
\caption{Ray-trace to free voxels}\label{alg:unk_algo}
\begin{algorithmic}[1]
\Function{RayTrace}{$ms\_grid, \boldsymbol{T}^v_c, x_i, y_i, z_i, vox\_size$}
\State $\boldsymbol{p}^v_c = \boldsymbol{T}^v_c [0 \ 0 \ 0 \ 1]^t$
\State $ray\_start \gets \boldsymbol{p}^v_c$
\State $ray\_dir \gets \boldsymbol{T}^v_c [x_i \ y_i \ z_i]^t$
\State $max\_dist \gets vox\_size\sqrt{x_i^2 + y_i^2 + z_i^2}$
\State $traversed\_dist \gets 0$
\State $vox\_val \gets free$
\While{$traversed\_dist < max\_dist$}
\State move by a voxel using \cite{amanatides1987fast} and set $x_i, y_i, z_i$ to \par\hskip\algorithmicindent new traced voxel
\If {$loc\_grid[x_i,y_i,z_i] == occupied$}
\State $vox\_val \gets unknown\ and\ traced$
\Else 
\State $ms\_grid[x_i,y_i,z_i] \gets vox\_val$
\EndIf
\State $traversed\_dist \gets vox\_size\sqrt{x_i^2 + y_i^2 + z_i^2}$
\EndWhile
\EndFunction
\end{algorithmic}
\end{algorithm}

\subsection{Update the local voxel grid}
After populating the measurement voxel grid with occupied voxels and ray-tracing to free voxels, the local voxel grid is updated with the \textbf{free}, \textbf{occupied}, and \textbf{unknown and traced} voxels, i.e. the previous values are replaced with the values of the \textbf{free}, \textbf{occupied},  and \textbf{unknown and traced} voxels of the measurement voxel grid  (Alg. \ref{alg:upd_algo}, line 2-4).

The grids are 1 dimensional arrays with the index $idx = x_i + y_i\times grid\_size_x + z_i\times grid\_size_x\times grid\_size_y$.

On the GPU the update is done in parallel resulting in a significant speed up. The speed up is due to being able to simultaneously access different memory addresses of the same array.

\begin{algorithm}
\caption{Update local grid with measurement grid}\label{alg:upd_algo}
\begin{algorithmic}[1]
\Function{UpdateGrid}{$loc\_grid, ms\_grid, idx$}
\If {$ms\_grid[idx] \ne unknown$}
\State $loc\_grid[idx] \gets ms\_grid[idx]$
\EndIf
\EndFunction
\end{algorithmic}
\end{algorithm}

\subsection{Shift the local voxel grid}
Every time the robot moves by a voxel or more, the local map is shifted to be centered at the new robot position
 and the new voxels resulting from that shift are initialized as \textbf{unknown}. This allows to always have the most relevant/close obstacle information to the robot. The shift is done in voxel units so we can simply use/copy voxels from the old local voxel grid (before shifting).

The shift is done after the local map is updated.
 This way it doesn't add any unnecessary latency. This step is solely done on the CPU.

\section{Simulation results} \label{sect:sim_res}
\begin{figure*}
\begin{subfigure}{0.33\textwidth}
\centering
\includegraphics[trim={0cm 0cm 0cm 0cm},clip,width=1\linewidth]{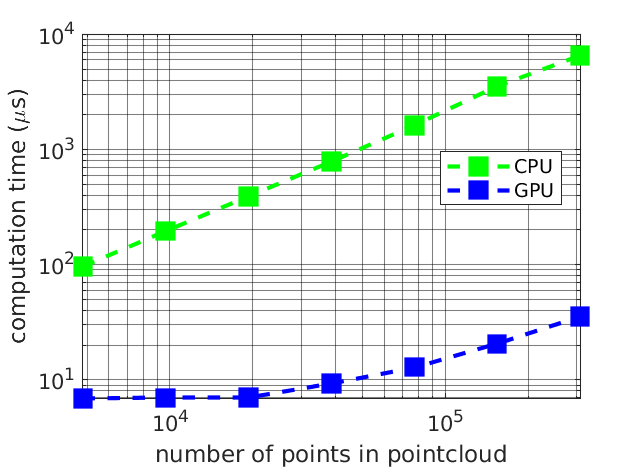}
\caption{A comparison between the CPU and GPU implementation of step 2 (populating occupied voxels) of our method.}
\label{fig:occ_time}
\end{subfigure}
\begin{subfigure}{0.33\textwidth}
\centering
\includegraphics[trim={0cm 0cm 0cm 0cm},clip,width=1\linewidth]{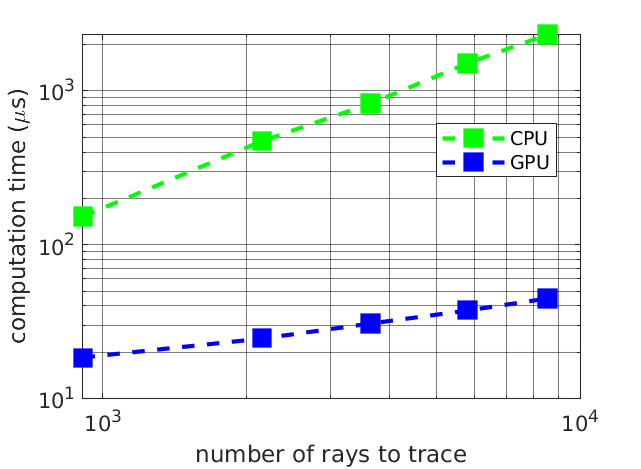}
\caption{A comparison between the CPU and GPU implementation of step 3 (ray-tracing to free voxels) of our method.}
\label{fig:unk_time}
\end{subfigure}
\begin{subfigure}{0.33\textwidth}
\centering
\includegraphics[trim={0cm 0cm 0cm 0cm},clip,width=1\linewidth]{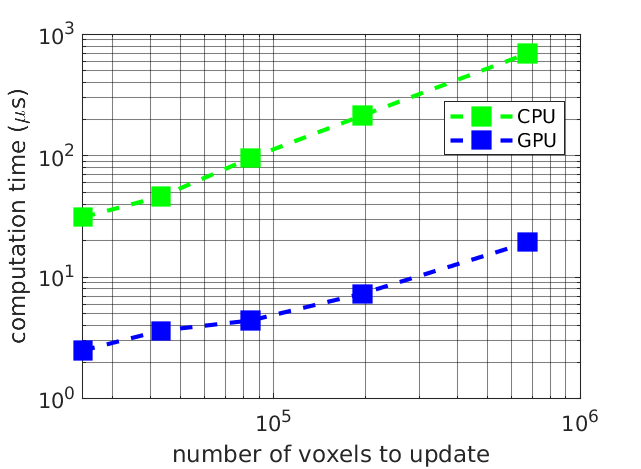}
\caption{A comparison between the CPU and GPU implementation of step 4 (merging the measurement grid with the local grid) of our method.}
\label{fig:upd_time}
\end{subfigure}
\caption{A comparison between the CPU and GPU implementation of steps 2-4. The CPU implementation is sequential and runs on a single core.}
\end{figure*}

We run the simulation on the following hardware setup: for the CPU we use Intel Core i7-9750H up to 4.50 GHz, and for the GPU we use NVIDIA's GeForce RTX 2060 up to 1.62 GHz.

We simulate the robot using mit-acl-gazebo\footnote{https://gitlab.com/mit-acl/lab/acl-gazebo}. The sensor is an RGB-D camera that outputs a depth image with a resolution of $320\times240$. The maximum $depth$ is 6.5 meters, $fov_x = 85 \deg$ and $fov_y = 101 \deg$. The chosen size of the voxel grid is $(grid\_size_x = 15 \ m, grid\_size_y = 15 \ m, grid\_size_z = 3 \ m)$
 and the voxel size $vox\_size = 0.15 \ m$. The robot size is $0.3 \ m$ which implies $vox\_inf = 2$.
  This results in $200\,000$ voxels, and $vox\_height\times vox\_width \approx 8\,500$ rays to trace. These are the standard parameters that we use for the simulation time comparisons unless specified otherwise.

We compare the performance of every step run on the CPU with its GPU version.
 We analyse the efficiency of our GPU implementation in terms of warp state statistics and warp cycles per execution
  instruction.
  The occupancy metric of the GPU
  is not studied as it depends on other factors than the efficiency of the algorithm such as the number of blocks and threads resulting from the pointcloud size/number of rays to trace/number of voxels in a grid. We choose 128 threads per block (4 warps).

\subsection{Populate occupied voxels - simulation}
We compare the computation time of the CPU and GPU implementation of the functional block "populate occupied voxels"
 as the number of obstacle points increases (Fig. \ref{fig:occ_time}). The scale of the $x$ and $y$ axes is logarithmic. 
The GPU computation time starts quasi-constant then increases linearly as the number of points surpasses $20\,000$. This is due to the GPU not being fully occupied before reaching the inflexion point. The CPU computation time  scales linearly with the number of points in the pointcloud. This is due to the fact that the time complexity of processing one point is constant. 

The GPU computation time slope in the linear segment is smaller then that of the CPU. This is expected due to the SIMT architecture and the efficiency of our implementation.

The GPU implementation outperforms the CPU implementation $14\times$ for $5\,000$ points
 and the difference grows bigger as the number of obstacle points increases. For $300\,000$ point the difference in performance is $185\times$.

\subsubsection*{GPU performance}
The average active threads per warp (with predication) is $32/32$, and the average not predicated off threads per warp is $29.65/32$. The average warp cycles per executed instruction (which defines the latency between two consecutive instructions) is 11.85.
This shows that our implementation for this step is efficient on the GPU.

\begin{figure*}[]
\begin{subfigure}{0.245\textwidth}
\centering
\includegraphics[trim={0cm 0cm 0cm -1cm},clip,width=0.9\linewidth]{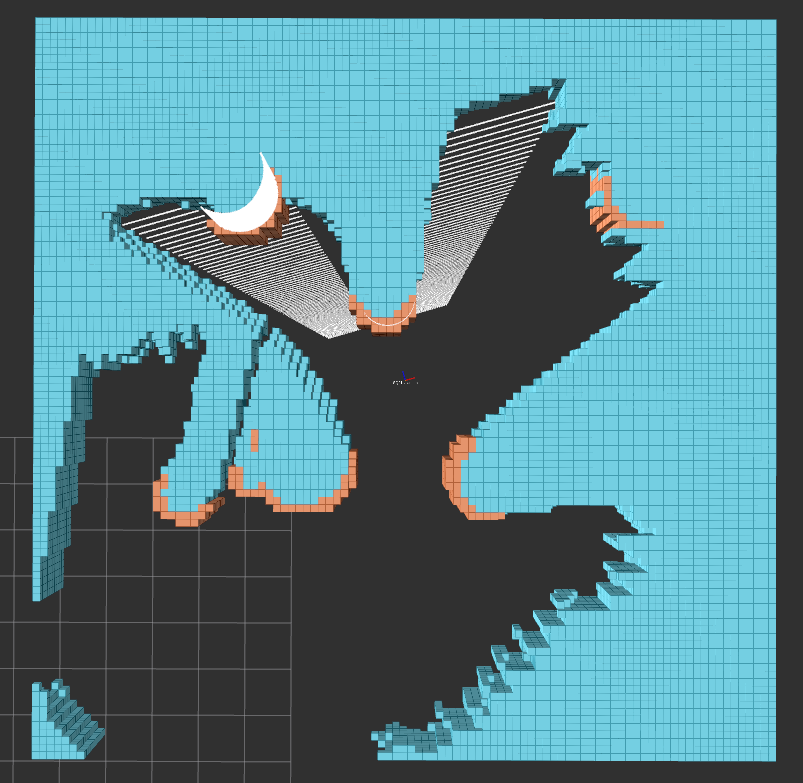}
\caption{case 1: our method}
\end{subfigure}
\begin{subfigure}{0.245\textwidth}
\centering
\includegraphics[trim={0cm 0cm 0cm -1cm},clip,width=0.87\linewidth]{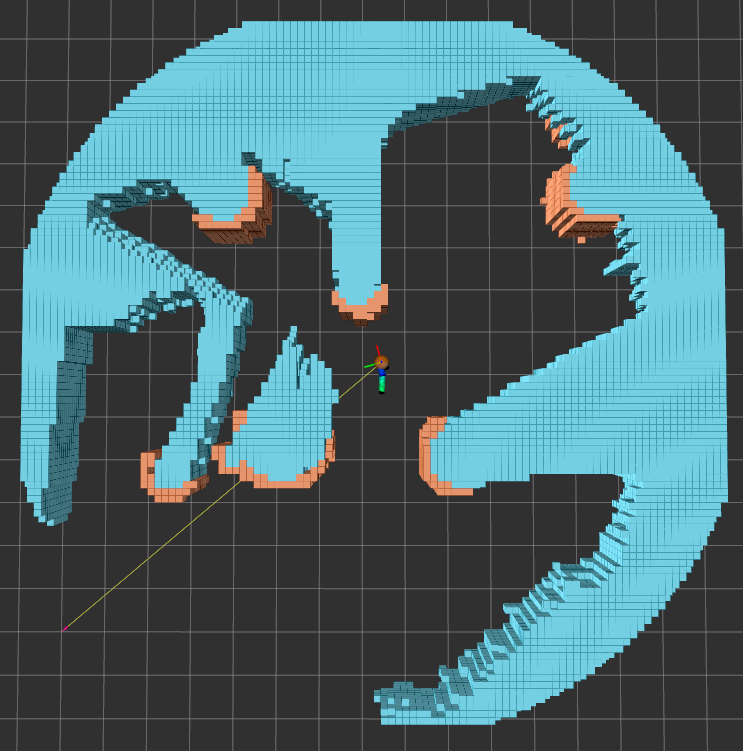}
\caption{case 1: mit-acl-mapping}
\end{subfigure}
\begin{subfigure}{0.245\textwidth}
\centering
\includegraphics[trim={0cm 0cm 0cm -1cm},clip,width=0.9\linewidth]{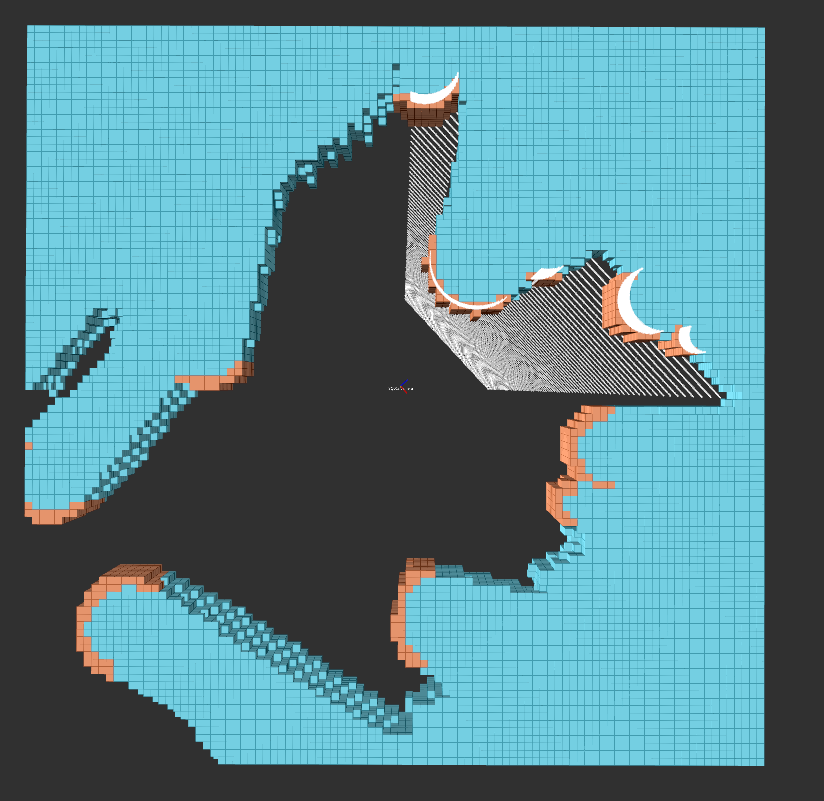}
\caption{case 2: our method}
\end{subfigure}
\begin{subfigure}{0.245\textwidth}
\centering
\includegraphics[trim={0cm 0cm 0cm -1cm},clip,width=0.87\linewidth]{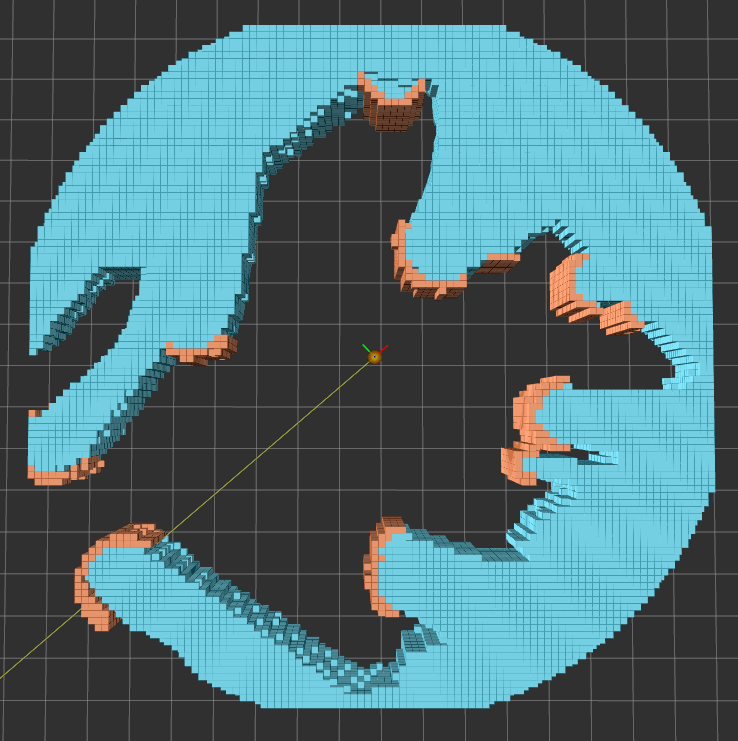}
\caption{case 2: mit-acl-mapping}
\end{subfigure}

\caption{We show the voxel grid generated by our method and mit-acl-mapping during a MAV exploration task using \cite{tordesillas2020faster}. The occupied voxels are shown in \textbf{orange}, the unknown voxels in \textbf{blue}, and the free voxels are transparent. The two methods deliver close results.}
\label{fig:cases}
\end{figure*}

\begin{table*}[ht]
\centering
\begin{tabular}{c | cccc | c}\hline
            & \begin{tabular}{@{}c@{}}Populate \\ occupied voxels\end{tabular} & \begin{tabular}{@{}c@{}}Ray-trace \\ to free voxels\end{tabular} & \begin{tabular}{@{}c@{}}Update local grid \\ with measurement grid\end{tabular}  & \begin{tabular}{@{}c@{}}CUDA \\ memory operations\end{tabular} & Total \\ \hhline{======} 
CPU   &    $0.1 \ ms$  & $2.2 \ ms$ &  $0.2 \ ms$    & - & $2.5\ ms$   \\ \hline
GPU  & $0.007 \ ms$   & $0.044 \ ms$ & $0.007 \ ms$  & $0.712 \ ms$ & $0.77\ ms$ \\ \hhline{======} 
Improvement &  $14.2 \times$  &  $50 \times$  &    $28.5 \times$   &  -  &    $3.3 \times$ \\ \hline
\end{tabular}
\caption{CPU and GPU average computation time comparison for the different functional blocks using the standard parameters (detailed in Sect. \ref{sect:sim_res}) during an exploration task. The voxel grid setup is not included (negligible time) and the local grid shifting is not included (done only on the CPU after the local map is updated - doesn't affect latency).}
\label{table:comp_time}
\end{table*}

\subsection{Ray-tracing to free voxels - simulation}
We compare the computation time of the CPU and GPU implementation of the functional block "ray-tracing to free voxels" as the number of rays to trace increases (Fig. \ref{fig:unk_time}). The scale of the $x$ and $y$ axes is logarithmic. 

Both CPU and GPU computation times scale linearly. This is expected as the computation time of a single ray-tracing function is limited by $max\_dist$ and thus has a constant upper bound. The slope of the GPU computation time is also smaller then that of the CPU in this case.
The GPU implementation outperforms the CPU implementation $8\times$ for $1\,000$ rays and the difference grows bigger as the number of rays to trace increases. For $8\,500$ rays the difference in performance is $52\times$.

\subsubsection*{GPU performance}
The average active threads per warp (with predication) is $24.12/32$, and the average not predicated off threads per warp is $20.57/32$. The reason is that we use the method described in \cite{amanatides1987fast} for ray tracing, which includes conditional if-else statements that cause branching. Branching in general degrades GPU performance and leads to less active threads per warp and wasted cycles. The average warp cycles per executed instruction is $11.87$.

\begin{figure}
\centering
\includegraphics[trim={0cm 0cm 0cm 0cm},clip,width=0.95\linewidth]{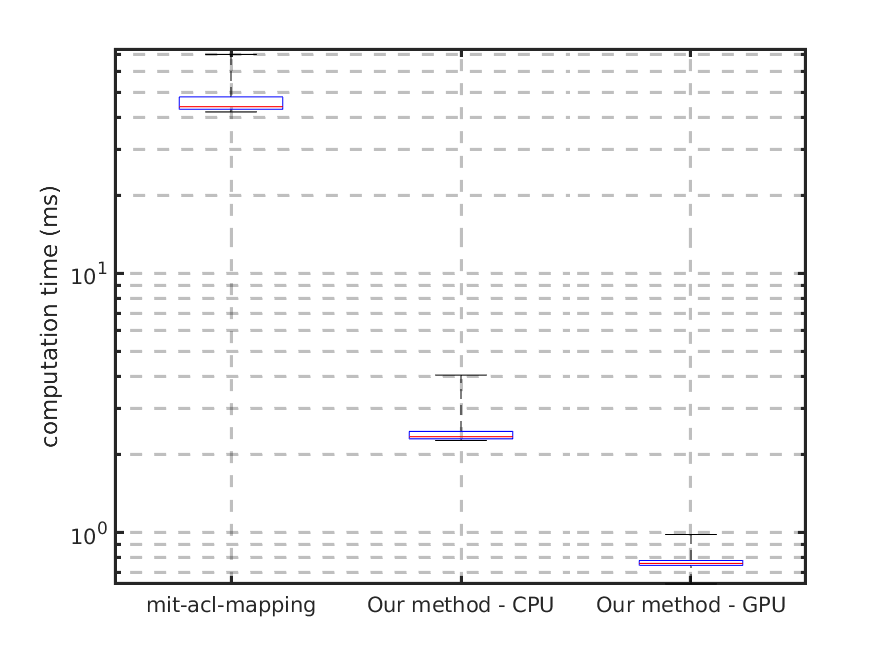}
\caption{A comparison between mit-acl-mapping \cite{ryll2019efficient}, the CPU and GPU implementation of our method. The red line represents the median. The lower and upper bounds of the box represent the 25\textsuperscript{th} and 75\textsuperscript{th} percentile respectively, and the lower and upper whiskers represent the minimum and maximum respectively.}
\label{fig:comp_time}
\end{figure}

\subsection{Update the local voxel grid - simulation}
We compare the computation time of the CPU and GPU implementation of the functional block "update the local voxel grid" as the number of voxels in the grid increases (Fig. \ref{fig:unk_time}). The scale of the $x$ and $y$ axes is logarithmic. 

Both CPU and GPU computation times scale linearly. This is expected as the computation time is constant (read/write access). The slope of the GPU computation time is also smaller then that of the CPU in this case because of the concurrent memory access operations that the GPU is capable of. The GPU implementation outperforms the CPU implementation $12.5\times$ for $24\,000$ voxels and the difference grows bigger as the number of voxels increases. For $675\,000$ voxels the difference in performance is $38\times$.

\subsubsection*{GPU performance}
The average active threads per warp (with predication) is $29.81/32$, and the average not predicated off threads per warp is $29.18/32$. It increase slightly as the number of voxels in a grid increases. The active warps is not $32/32$ due to the \textit{if} statement that verifies that a voxel of the measurement voxel grid is not \textbf{unknown} before merging it with the local voxel grid. The average warp cycles per executed instruction is $46.75$.

\subsection{Comparison with mit-acl-mapping}
The proposed method vastly outperforms mit-acl-mapping in terms of computation time (Fig. \ref{fig:comp_time}) while delivering similar results (Fig. \ref{fig:cases}). 
 The reported computation times are for the setup explained at the beginning of Section \ref{sect:sim_res}, while doing an exploration task using \cite{tordesillas2020faster}. We show the distribution of the computation time (Fig. \ref{fig:comp_time}) as a boxplot.
The CPU implementation of our method takes on average $2.5\ ms$ and is $18\times$ faster then mit-acl-mapping. The GPU implementation takes on average $0.77\ ms$ and is $3.3\times$ faster than the CPU implementation. 
Note that in addition to the kernels' execution time (steps 2,3 and 4), the GPU time includes allocating (\textit{cudaMalloc}) and freeing (\textit{cudaFree}) memory on the device (GPU), setting this memory (\textit{cudaMemset}) and transferring data from the host (CPU) to the device and from the device to the host (\textit{cudaMemcpy}). On average, \textit{cudaMalloc} takes $308 \ \mu s$, \textit{cudaMemcpy} takes $286 \ \mu s$, \textit{cudaFree} takes $97 \ \mu s$, \textit{cudaMemset} takes $21 \ \mu s$ and the kernels (launch + execution) take $58 \ \mu s$ (Table \ref{table:comp_time}).

The results (Fig. \ref{fig:cases}) show that our method delivers close results to mit-acl-mapping but more conservative (less voxels are freed) which is expected since our ray is stopped when it hits an occupied voxel whereas mit-acl-mapping uses \cite{bresenham1965algorithm} to trace every pixel of the image (Fig. \ref{fig:simple_ray}). The method used by mit-acl-mapping cannot be implemented on a GPU as is, and the ray-tracing method used \cite{bresenham1965algorithm} can miss voxels that the ray passes through, unlike \cite{amanatides1987fast} which we use.

\section{Conclusions and Future Works}
We presented a novel method for the generation of voxel grids with occupied, free and unknown voxels. The method is efficient while sacrificing little accuracy. We compared our method to the state-of-the-art and implemented a GPU version of it which resulted in a considerable speed up. The CPU and GPU implementations outperform the state-of-the-art in computation time while delivering similar quality.


\bibliographystyle{IEEEtran}
\bibliography{IEEEabrv,IEEEexample}

\end{document}